\documentclass[11pt]{article}

\usepackage[margin=1in]{geometry}
\usepackage{lmodern}        
\usepackage[T1]{fontenc}
\usepackage{amsmath,amssymb}
\usepackage{graphicx}
\usepackage{booktabs}
\usepackage{array}
\usepackage{microtype}
\usepackage{xcolor}
\usepackage{authblk}
\usepackage[numbers,sort&compress]{natbib}
\usepackage[colorlinks=true,linkcolor=blue,citecolor=blue,urlcolor=blue]{hyperref}
\usepackage{titlesec}
\titleformat{\section}{\large\bfseries}{\thesection}{0.6em}{}
\titleformat{\subsection}{\normalsize\bfseries}{\thesubsection}{0.6em}{}

\newcommand{\C}{C}
\newcommand{\kk}{k}

\title{\bfseries Coverage, Not Targeting:\\ A Structural Regime in Multi-Turn Agent Credit Assignment}
\author[1,$\dagger$]{Chenyu Zhou}
\author[2]{Qiliang Jiang}
\author[3]{Shuning Wu}
\author[3,$\dagger$]{Xu Zhou}
\affil[1]{School of Engineering, Institute of Science Tokyo, Japan}
\affil[2]{College of Control Science and Engineering, Zhejiang University, China}
\affil[3]{Department of Electrical and Computer Engineering, National University of Singapore, Singapore}
\affil[$\dagger$]{Corresponding authors: \texttt{zhou.c.76d6@m.isct.ac.jp}, \texttt{zhouxu\_nus@u.nus.edu}.}
\date{}

\begin{document}
\maketitle

\begin{abstract}
Multi-turn agentic RL increasingly treats credit assignment as a \emph{targeting} problem: given a single
terminal verifiable reward, a fast-growing line of methods---turn-level reward models, learned per-step
calibration, process supervision---aims to localize credit onto the turns that ``really'' mattered. We identify the
structural quantity that predicts when this is the right move---the \textbf{verifier information density}
$V_d{=}\kk/\C$, the fraction of an agent's $\C$-step causal chain whose per-turn correctness the verifier exposes---
and show that terminal-state verifiers, which dominate current agentic-RL evaluation, sit deep in a low-$V_d$
regime where targeting is the wrong axis. There, in controlled shared-rollout comparisons on $\tau^2$-bench \citep{tau2} that
separate reward density from credit geometry, the harmful axis is not \emph{where} credit is targeted but
\emph{how much} it is concentrated: a continuous dense reward spread \emph{uniformly} beats the sparse binary
outcome reward---which is \emph{net-harmful}, degrading the base policy on $4/5$ seeds---while redistributing the
\emph{same} advantage onto ``progress'' turns, or onto \emph{random} turns, is equally harmful. Targeting is
second-order. The mechanism is \textbf{coverage}: terminal-state verification collapses the observable credit
signal to a single final-write turn ($\kk{=}1$ in $98\%$ of rollouts) while success requires a $5$--$8$ step chain
of hard-prerequisite tool calls, so any fixed-budget concentration under-covers it. A controlled synthetic
\textbf{phase boundary} (concentration wins only above a high crossover $V_d^\ast\!\approx\!0.8$) and a
real-rollout diagnostic (measured $\kk{=}1$, $\C{=}5$--$8$, so $V_d\!\approx\!0.15$) converge on this account,
which predicts uniform should also win on a second benchmark whose natively per-turn checker still exposes under
half the chain ($V_d\!\approx\!0.4$)---and it does, with the same concentration-not-targeting structure (shuffled
$8/8$ seeds, per-turn $6/8$). Uniform redistribution is thus the zero-information coverage default that per-turn schemes must beat, and
we contribute the matched-concentration \textbf{shuffled control} that any targeting claim should clear. The
credit-geometry effect holds within Qwen3 (8B and 14B) and reproduces \emph{across model families} on a
tool-specialized Llama-3.1 agent (ToolACE-2-8B; $\Delta{=}-0.048$ over $32$ pre-registered seeds, with an independent $20$-seed replication itself
significant at $\Delta{=}-0.054$). A pre-registered matched-budget breadth sweep on the same agent traces a monotone
dose--response whose deficit vanishes only at full chain coverage, and a reward-to-go arm reaches full-coverage parity,
as the account predicts. The structural ingredients ($\kk{=}1$, $\C{=}5$--$8$) are measured across both families.
\end{abstract}

\section{Introduction}
\label{sec:intro}

Reinforcement learning on multi-turn tool-using agents is increasingly trained against \textbf{terminal
verifiable rewards}: a program checks the final environment state (a database, a filesystem, an API ledger) and
returns a scalar at the end of the rollout. This poses a credit-assignment question that single-turn RLHF never
had to answer---a successful trajectory is a \emph{chain} of tool calls, and the verifier scores only its
endpoint. How should that endpoint signal be distributed across the turns that produced it?

Credit assignment is the central open problem of this setting, and the field's prevailing answer is to
\emph{add information}: turn-level reward models and critics \citep{mtgrpo,archer,sweetrl,proxmo}, learned per-step
calibration \citep{irc}, process reward models \citep{lightman}, hindsight relabeling \citep{her}---a fast-growing line of
work whose implicit premise is that better \emph{localization} of credit (putting reward where the ``real''
progress happened) is what is missing.

We report a result that inverts this premise. With \textbf{no new information}---redistributing the \emph{same}
terminal advantage---uniform spreading beats heuristic concentration, and the binary outcome reward is actively
harmful. The harmful axis is not \emph{where} credit is targeted but \emph{how concentrated} it is:
progress-targeted and random-targeted concentration are equally bad, while uniform spreading wins. Localization is
not the lever. That dense rewards can beat sparse ones \citep{toolrl,hybridrl}, and that naive per-turn rewards can
degrade performance \citep{irc}, has been reported before; our contribution is to isolate the \emph{responsible}
axis---concentration, not targeting---with a shuffled-credit control, and to explain it with a coverage mechanism
measured on real rollouts.

\begin{figure}[t]
\centering
\includegraphics[width=\textwidth]{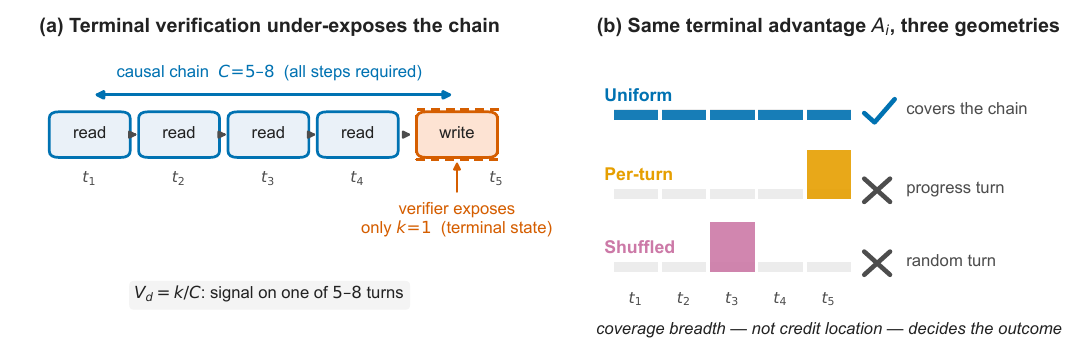}
\caption{\textbf{Why coverage, not location, governs credit assignment under terminal verification.} (a) A successful
rollout is a $\C=5$--$8$ chain of hard-prerequisite tool calls, but a state-transition verifier scores only the terminal
write, exposing a single turn ($\kk=1$, so $V_d=\kk/\C$ is $0.12$--$0.20$). (b) The same group-normalized terminal
advantage $A_i$, redistributed three ways: \emph{uniform} spreads it over every turn and covers the whole chain, whereas
\emph{per-turn} (progress turns) and \emph{shuffled} (random turns) both concentrate it on a single turn and under-cover
the chain. Because a lone spike lands on one of $5$--$8$ causal turns wherever it goes, coverage breadth---not credit
location---decides the outcome.}
\label{fig:overview}
\end{figure}

The explanation is \textbf{coverage}. Under terminal-state verification, the only turn whose observable progress
signal moves is the final batched write---so any progress-derived per-turn signal collapses to a single-turn
spike (we measure $\kk=1$ in $98\%$ of real rollouts). But task success depends on a $5$--$8$ step chain of
\emph{hard prerequisite} tool calls---look up the user, retrieve the order, find the replacement product, then
write the exchange; you cannot exchange an order you never looked up. Concentration on a fixed
budget of turns covers only a fraction of that chain and starves the rest; uniform spreading covers all of it
(Fig.~\ref{fig:overview}). When the causal chain is longer than the concentration budget ($\C>\kk$), uniform wins; this is a \textbf{phase
boundary}. We establish the boundary in a controlled synthetic environment (\S\ref{sec:toy}); the real model does
not cross it---it is measured to sit deep on the $\C\gg\kk$ side (\S\ref{sec:diag}). And it \emph{cannot} cheaply
cross it: absent oracle labels or a learned judge, a state-transition verifier observes only steps that change
state, so $V_d$ is upper-bounded by the task's \emph{write-fraction} and read/exploration steps are a structural
blind spot---a census of common multi-turn tool-agent benchmarks finds none escaping this ceiling (the highest, BFCL~V3's per-turn
checker, reaches only $V_d\!\approx\!0.4$; \S\ref{sec:writefrac}). Uniform coverage is thus not a tuning choice but
the structurally forced default for terminal-verifier agents.

\paragraph{Contributions.}
\begin{enumerate}\itemsep2pt
\item \emph{The verifier information density $V_d{=}\kk/\C$ as a structural regime map for credit geometry.}
$V_d$---the fraction of an agent's $\C$-step causal chain whose per-turn correctness the verifier exposes---is both
\emph{measurable} (from real rollouts) and \emph{predictive}: a controlled synthetic \textbf{phase boundary} shows
fixed-budget concentration pays off only above a high crossover ($V_d^\ast\!\approx\!0.8$), while terminal-state
verifiers, which dominate agentic-RL evaluation, are \emph{measured} to sit at low $V_d\!\approx\!0.15$
($\kk{=}1$, $\C{=}5$--$8$)---deep in the coverage regime (\S\ref{sec:cov}, \S\ref{sec:vd}).
\item \emph{A structural ceiling on $V_d$: the write-fraction bound.} A state-transition verifier can score a step
only if the step changes state, so---absent oracle goal-labels or a learned judge---the independently checkable
fraction of the causal chain is upper-bounded by the task's \emph{write-fraction}; read/exploration steps, whose
correctness is goal-relative, are a permanent structural blind spot. A census of common multi-turn tool-agent
benchmarks (Table~\ref{tab:census}) finds none that exposes a naturally high $V_d$: the highest, BFCL~V3's
per-turn checker, reaches only $V_d\!\approx\!0.4$, and the terminal-verifier majority sits at
$V_d\!\approx\!0.1$--$0.2$. This explains \emph{why}
the coverage regime is the common case---a structural property of state-based verification, not an accident of our
tasks---and recasts learned or oracle process rewards as paying to fill this observability gap rather than acting as
free streaming verifiers (\S\ref{sec:writefrac}).
\item In that low-$V_d$ regime, a controlled isolation (shared-rollout) shows dense-uniform $>$ binary (binary
\emph{net-harmful}) and per-turn concentration harmful---only the credit geometry differs across arms
(\S\ref{sec:density}). And it is \emph{concentration, not targeting}: random-targeted $\approx$ progress-targeted
$\ll$ uniform, so targeting is second-order and coverage breadth first-order (\S\ref{sec:targeting}). The
\textbf{coverage} mechanism explains why---observable credit collapses to $\kk{=}1$ while success needs the
$5$--$8$ step chain---and is algorithm-agnostic (GRPO, advantage-baseline, REINFORCE at matched effective learning
rate) (\S\ref{sec:cov}).
\item \emph{A methodological control for targeting claims.} Because changing credit geometry alters concentration
and optimization dynamics \emph{together}, isolating a genuine \emph{targeting} benefit requires a
matched-concentration \textbf{shuffled control} that holds coverage breadth fixed and varies only \emph{which}
turns are credited. We use this control throughout and recommend it as a default guard for per-turn
credit-assignment claims (\S\ref{sec:targeting}, \S\ref{sec:limits}).
\item Boundary conditions that unify the null regimes under the same mechanism (a capability floor, a
resolution/partial-density gate), and reproduction of the effect on a second benchmark (BFCL~V3, $V_d\!\approx\!0.4$)
and \emph{across model families} (a tool-specialized Llama-3.1 agent), which also passes a pre-registered
matched-budget breadth sweep: at fixed total credit, performance rises monotonically with chain coverage, and a companion
reward-to-go arm on Qwen3-8B reaches full-coverage parity (\S\ref{sec:boundary}). Uniform redistribution
is the strong zero-information default for the $\C\gg\kk$ regime, the common case for terminal-verifier tool agents.
\end{enumerate}

\paragraph{Roadmap.} We first isolate the empirical effect at the measured low-$V_d$ operating point
(\S\ref{sec:density}--\S\ref{sec:targeting}), then lift it into the verifier-density phase diagram
(\S\ref{sec:cov}--\S\ref{sec:vd}) and test its predictions across capability, model family, credit breadth, and a
second benchmark (\S\ref{sec:boundary}).

\section{Setup}
\label{sec:setup}

\paragraph{Environment.} $\tau^2$-bench \texttt{retail}, restricted to its pure database-state tasks (the verifier
checks DB field-level changes; no natural-language-judged assertions). Base agent Qwen3-14B (we also report 8B
and 4B for the scale analysis). Training is GRPO \citep{grpo} with LoRA \citep{lora} (rank 16; full
configuration in Appendix~\ref{app:config}). BFCL~V3 multi-turn \citep{bfcl} provides a second benchmark, and a
second $\tau^2$-bench domain (telecom) a partial-progress contrast for the resolution gate (\S\ref{sec:boundary}).

\paragraph{Shared-rollout isolation (the clean comparison).} For each seed we collect \emph{one} batch of
base-policy rollouts and train every arm from the \emph{same} batch with the same optimizer. Two axes are
separated. Along the \emph{density} axis, the binary arm scores each rollout with the coarse $0/1$ outcome while
the dense arms use the continuous DB-field progress. Along the \emph{geometry} axis, the dense arms share a single
group-normalized terminal advantage $A_i$ and differ \emph{only} in how it is distributed across turns. The
geometry comparison thus isolates credit distribution from every confound (data, exploration, terminal signal,
magnitude). This isolation exists \emph{only} at the first policy update: once training iterates, each arm
collects its own data, and any downstream difference conflates credit geometry with data collection. The
single-update comparison is therefore the object of study.

\paragraph{Arms.} We compare four credit geometries.\footnote{Released-harness names: Binary~=~\texttt{official},
Uniform~=~\texttt{dense\_full}, Per-turn~=~\texttt{dense\_perturn}, Shuffled~=~\texttt{dense\_perturn\_shuf}.}
\textbf{Binary}---the conventional sparse $0/1$ outcome reward. \textbf{Uniform}---a continuous dense reward
(DB-field progress) with the group-normalized advantage assigned to every turn, $a_t=A_i$ (the standard GRPO
allocation). \textbf{Per-turn}---the same $A_i$ redistributed by each turn's
share of positive DB-field progress $\Delta S_t$, so that $\sum_t a_t=A_i$ (progress-targeted concentration). \textbf{Shuffled}---the
per-turn weights permuted to \emph{random} turns (random-targeted concentration; same concentration, no targeting
information). A global gradient-norm clip of $1.0$, saturated in every logged epoch of every arm, equalizes the
applied update magnitude across geometries, which therefore differ in update direction, not size. Note that the per-turn arm consumes strictly more verifier information than uniform---it sees
\emph{where} the progress occurred---so the geometry comparison is, if anything, biased toward per-turn.

\paragraph{Metric.} Held-out environment-assertion support on the pre-registered gradable subset (primary,
continuous) and official task success (co-primary, binary); the formal definition, the frozen gradable task list, and
the user-simulator protocol are in Appendix~\ref{app:metric}. We report per-seed paired deltas (sign test) and a
one-sided per-task Wilcoxon signed-rank test in the pre-registered direction on the held-out gradable subset (95
task--seed pairs for the main comparison).
All confirmatory arm comparisons are \textbf{within-run paired}---competing geometries share each seed's
training rollouts and evaluation batch; Appendix~\ref{app:evalvar} quantifies the between-batch evaluation
variability that makes unpaired cross-run comparisons unsafe at this evaluation budget.

\section{Density is necessary but not sufficient}
\label{sec:density}

\begin{figure}[t]
\centering
\includegraphics[width=0.7\textwidth]{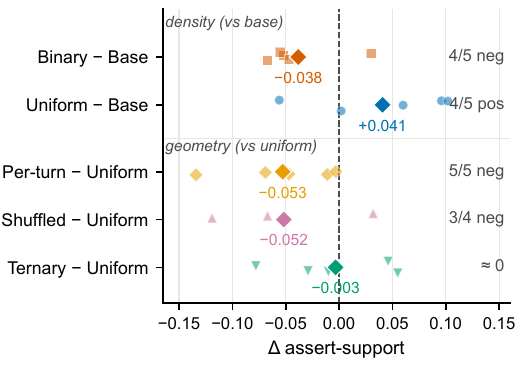}
\caption{\textbf{Credit geometry as paired per-seed contrasts} (retail-14B, held-out assert-support; faint markers are
per-seed differences, diamonds means). Binary degrades the untrained base while dense uniform improves it; relative to
uniform, concentrated per-turn and shuffled are harmful while ternary is indistinguishable. Inference is task-level
(one-sided Wilcoxon over $95$ task--seed pairs). Concentration is the harm; targeting is second-order.}
\label{fig:results}
\end{figure}

\paragraph{Uniform dense beats binary, and binary is net-harmful.} The uniform arm exceeds the binary arm by
$+0.079$ on gradable assert-support (sign $4/5$ across seeds; one-sided per-task Wilcoxon $p=0.003$ over the $95$
task--seed pairs; seed-then-task hierarchical bootstrap $95\%$ CI $[+0.009,+0.149]$, Appendix~\ref{app:evalvar}), and
by $+5.2$pp on official success. The separation is \emph{twofold}: relative to the
untrained base, the binary reward \emph{degrades} the policy ($4/5$ seeds) while the uniform dense reward
\emph{improves} it ($4/5$ seeds; $+0.041$ over base, Table~\ref{tab:main}). The conventional sparse outcome reward
is not merely weaker---under matched conditions it actively hurts (Fig.~\ref{fig:results}).

\paragraph{Resolution is cheap.} A ternary discretization of the dense reward (three levels: $0$, $0.5$, $1$)
retains $96\%$ of the advantage ($+0.076$ over binary, $5/5$): the gain comes from having \emph{any}
non-degenerate gradient on groups where every rollout otherwise fails, not from fine-grained reward resolution.

\paragraph{But concentrating the dense reward hurts.} Redistributing the \emph{same} terminal advantage onto
progress turns (the per-turn arm) underperforms uniform on every seed on assert-support (sign $0/5$, mean $-0.053$)
and on official task success (mean $-0.063$, $4/5$ seeds; per-seed values in Appendix~\ref{app:diag}), a deficit
that persists across the entire trainable dose range (Appendix~\ref{app:dose}) and reproduces in the powered
same-batch four-arm comparison on BFCL~V3 ($-0.036$, paired $t=-2.80$; \S\ref{sec:boundary}). Density is
necessary; how the density is distributed decides whether it helps (Fig.~\ref{fig:results}). The matched-budget sweep
of \S\ref{sec:boundary} reproduces this deficit at strictly equal total credit and separates its coverage and scale
components on a second benchmark.

\begin{table}[t]
\centering
\caption{Held-out gradable assert-support, per seed (retail-14B). Uniform is the reference for $\Delta$. Binary
degrades the base while uniform improves it (twofold separation). Per-turn and shuffled (both concentrated) are
below uniform; ternary $\approx$ uniform. Shuffled is over four seeds; its mean and deltas
are computed on those four, where the per-turn mean is $0.499$ (hence the $+0.014$ shuffled$-$per-turn difference
in \S\ref{sec:targeting}, not $0.513-0.510$). On official success the binary--uniform gap is $-5.2$pp.}
\label{tab:main}
\small
\begin{tabular}{lcccccccc}
\toprule
arm & s1 & s2 & s3 & s4 & s5 & mean & $\Delta$ base & $\Delta$ unif. \\
\midrule
base (untrained) & 0.597 & 0.502 & 0.509 & 0.506 & 0.496 & 0.522 & $+0.000$ & $-0.041$ \\
binary outcome   & 0.545 & 0.447 & 0.539 & 0.439 & 0.449 & 0.484 & $-0.038$ & $-0.079$ \\
dense uniform    & 0.541 & 0.504 & 0.605 & 0.608 & 0.556 & \textbf{0.563} & $+0.041$ & $+0.000$ \\
dense per-turn   & 0.494 & 0.493 & 0.471 & 0.539 & 0.553 & 0.510 & $-0.012$ & $-0.053$ \\
dense shuffled   & 0.488 & 0.536 & 0.538 & 0.489 & --    & 0.513 & $-0.016$ & $-0.052$ \\
dense ternary    & 0.587 & 0.475 & 0.595 & 0.530 & 0.611 & 0.560 & $+0.038$ & $-0.003$ \\
\bottomrule
\end{tabular}
\end{table}

\section{It is concentration, not targeting}
\label{sec:targeting}

If per-turn concentration loses because it targets the \emph{wrong} turns, then targeting \emph{random} turns
should be strictly worse. It is not. The shuffled arm (random-targeted) lands within noise of the per-turn arm
(progress-targeted)---both well below uniform (shuffled $-$ per-turn mean $+0.014$ on the four shuffled seeds,
$95\%$ CI $[-0.07,+0.10]$). This is a null on targeting, not a certified equivalence. But the direction is clear---moving the same
concentrated credit to random turns does not make it worse---and the coverage mechanism of \S\ref{sec:cov}
explains why: at $\C\gg\kk$ a single spike covers $\sim\!1$ of $5$--$8$ causal turns wherever it lands, so
\emph{where} it lands is second-order and \emph{that} it is concentrated is the harm. The first-order quantity is
coverage breadth. The shuffled arm removes exactly one thing---the information about which turns carried
progress---while holding the concentration budget fixed; that its removal costs nothing is direct evidence that the
per-turn deficit is not a mis-targeting failure that sharper localization could repair. Any scheme that concentrates a
fixed advantage budget, however well it chooses the target, therefore inherits the same coverage risk in this regime.

\paragraph{No targeting scheme isolates a robust benefit.} As a robustness check on whether \emph{any} plausible
targeting scheme can beat uniform once coverage is held fixed, a wider battery of matched-concentration,
shared-rollout credit geometries on Qwen3-8B ($4$ seeds) stress-tests the same conclusion: crediting verifier-confirmed gold-relevant read turns, a matched count of \emph{random} read turns,
write turns only, or gold reads only. No
variant isolates a robust benefit from \emph{which} turns receive credit once coverage breadth is held fixed; we
report this as directional corroboration ($n{=}4$) of the shuffled-control result above, because these auxiliary arms
were trained and evaluated in separate runs (Appendix~\ref{app:evalvar}). The powered, within-run-paired version of
this control is the pre-registered cross-family breadth sweep of \S\ref{sec:boundary}, which reproduces the null.

\section{The coverage account}
\label{sec:cov}

\subsection{A synthetic phase boundary}
\label{sec:toy}
In a controlled multi-turn GRPO toy with \emph{known} causal structure ($\C$ of $T$ turns are causally necessary;
a concentration budget covers $\kk$ turns), we sweep causal sparsity $\C$. Uniform credit is robust across all
sparsities. Fixed-budget concentration wins only when it can cover the whole causal set ($\C\le\kk$) and loses
monotonically as the causal chain outgrows the budget ($\C>\kk$; Fig.~\ref{fig:coverage}b)---a prediction the
real-model breadth sweep confirms directly (\S\ref{sec:boundary}). The crossover at $\C\approx\kk$ is the
mechanism: concentration is a coverage bet that pays off only when the causal structure is sparse enough to fit
the budget. The boundary is robust to budget size, network capacity, graded (rather than binary) causality, and
sequential dependency, and persists under collect-from-policy training in which each arm collects from its own
evolving policy. It is \textbf{algorithm-agnostic}: once effective learning rate is matched, uniform wins
under GRPO, an advantage-baseline, and REINFORCE---raw cross-algorithm differences are an effective-LR confound,
not a GRPO artifact. The full environment specification is in
Appendix~\ref{app:toyspec}.

\subsection{A real-rollout coverage diagnostic}
\label{sec:diag}
\begin{figure}[t]
\centering
\includegraphics[width=\textwidth]{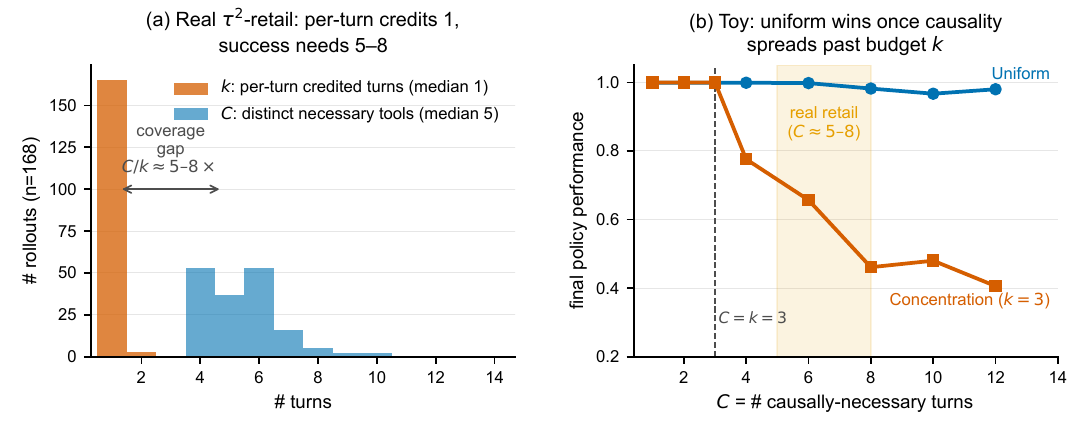}
\caption{\textbf{The coverage gap.} (a) Real $\tau^2$-bench rollouts: the per-turn credit budget is $\kk=1$
(spike on the terminal DB write) while the causal tool chain is $\C=5$--$8$; the prerequisite reads receive zero
per-turn credit. (b) The synthetic phase boundary: uniform wins iff $\C>\kk$; the real-retail operating point
($\kk=1$, $\C=5$--$8$) sits deep in the shaded $\C\gg\kk$ region.}
\label{fig:coverage}
\end{figure}

On real $\tau^2$-bench rollouts we measure the two quantities the toy predicts matter. The per-turn credit budget
---the participation ratio (effective number of turns carrying the credit signal) of the $\Delta S$ weights---is
$\kk=\text{median } 1.00$: in $98\%$ of rollouts the entire progress signal lands on a single turn, the terminal
batched DB write (the $2\%$ multi-turn cases do not move the regime). But success requires a chain of
$\C=5$--$8$ tool calls (median $8$ actions from $5$ distinct tools), whose first members are hard-prerequisite
reads---user lookup, order retrieval, product lookup---that change no DB state, and therefore receive \emph{zero}
per-turn credit, yet are physically necessary: an order cannot be exchanged without first retrieving it. This is a
hard \emph{data} dependency, not an inflated count: a static data-flow check confirms that in $100\%$ of
successful rollouts the terminal write's arguments (order, item, and product ids) are values returned by the
precursor reads (median $5$ reads precede the first write). The write cannot be constructed without them, and its
arguments cannot be assembled from the user instruction alone; the chain length is also insensitive to how it is
counted (Appendix~\ref{app:diag}). So $\C/\kk=5$--$8\times$:
the observable signal is maximally concentrated exactly where the causal structure is maximally distributed
(Fig.~\ref{fig:coverage}a).

\paragraph{This explains the targeting null.}
At $\C\gg\kk$, a single spike---whether on the progress turn (per-turn) or a random turn (shuffled)---covers
$\sim 1$ of $5$--$8$ causal turns. \emph{Which} turn it lands on is a second-order choice among
equally-incomplete coverings; only uniform covers the chain. This is precisely why shuffled $\approx$ per-turn
$\ll$ uniform.

\subsection{Verifier information density as the regime coordinate}
\label{sec:vd}
Is the collapse to $\kk=1$ intrinsic to multi-turn credit, or an artifact of the \emph{batched} terminal verifier?
We answer this in the controlled synthetic environment of \S\ref{sec:toy} by adding a second axis, the verifier information density $V_d$---the fraction
of the causal chain whose per-turn correctness the verifier exposes \emph{independently}. At $V_d\!\to\!0$ the
verifier reveals only a terminal aggregate, so any progress signal collapses to a single spike ($\kk=1$); at
$V_d\!=\!1$ every causal turn carries its own reward ($\kk\!\to\!\C$). Real terminal-verifier agents sit near the
low end: $\kk=1$ over a $\C=5$--$8$ chain is $V_d\!=\!\kk/\C\!\approx\!0.12$--$0.20$ (we use
$V_d\!\approx\!0.15$ as the representative value). Crucially, $V_d$ and $\C$
parameterize different subsystems---$V_d$ the verifier's observation model, $\C$ the task's causal structure; we
vary $V_d$ by verifier exposure at \emph{fixed} $\C$ (they couple through the budget $\kk$ in the toy, but a real
streaming verifier raises $V_d$ without changing $\C$). The two credit geometries are matched in total advantage
mass per rollout, so they differ only in shape.

\begin{figure}[t]
\centering
\includegraphics[width=\textwidth]{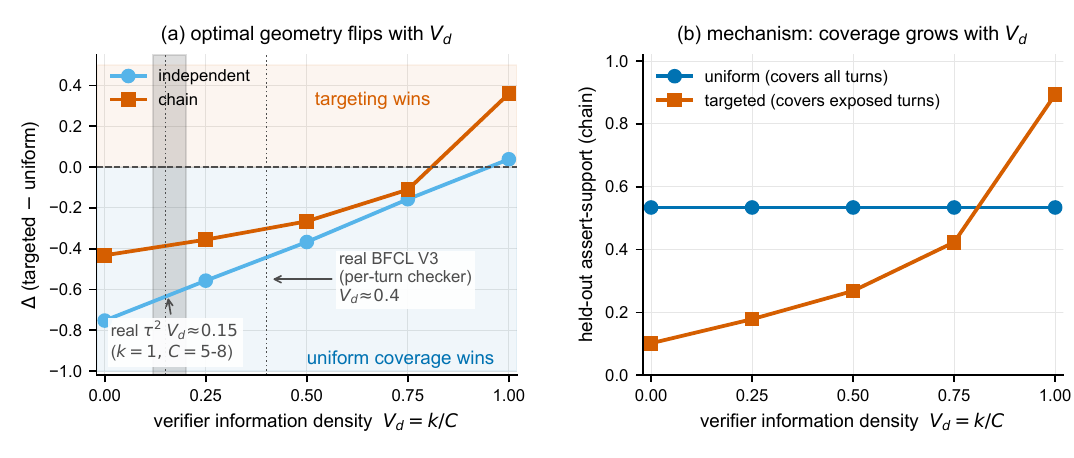}
\caption{\textbf{The relative performance of credit geometries is organized by verifier information density
$V_d$.} (a) Targeted$-$uniform held-out gap vs.\ $V_d$ (the fraction of the causal chain the verifier exposes
independently per turn); uniform coverage wins below the crossover $V_d^\ast$, targeting only above it. Real
terminal-verifier agents sit at $V_d\!\approx\!0.15$ ($\kk{=}1$, $\C{=}5$--$8$), deep in the coverage regime.
(b) Mechanism: the targeted arm's score rises purely as its coverage grows with $V_d$; uniform (full coverage) is
flat. ``Chain'' is the realistic state-dependent causal structure (the independent-turn structure, crossover
$\approx\!0.95$, is also shown); 32 paired seeds.}
\label{fig:vd}
\end{figure}

Sweeping $V_d$ traces a clean boundary (Fig.~\ref{fig:vd}). On the realistic state-dependent chain---a turn counts
only once all prerequisite turns are also correct---the targeted-minus-uniform gap closes monotonically with
$V_d$: $-0.43, -0.36, -0.27, -0.11, +0.36$ across $V_d=0\ldots1$ (32 paired seeds), crossing zero only at
$V_d^\ast\!\approx\!0.81$. Uniform coverage wins until the verifier exposes roughly four-fifths of the chain. This
crossover is not an artifact of one chain length: sweeping $\C=4$--$10$, $V_d^\ast$ stays in $[0.75, 0.91]$---in
every case far above where real terminal-verifier agents operate ($V_d\!\approx\!0.15$). The same coverage budget
is at work as in \S\ref{sec:toy}---below $V_d^\ast$ the verifier exposes too little of the chain for concentration
to cover it---but now turned by the verifier's observation model rather than by chain length, the two being
independent knobs. (On independent-turn chains, where uniform is near its ceiling, the crossover comes later,
$V_d^\ast\!\approx\!0.95$; Fig.~\ref{fig:vd}a.) The mechanism is the one
from \S\ref{sec:toy}: at low $V_d$ the targeted arm can only cover the few turns the verifier exposes (the rest
receive zero gradient) while uniform covers the whole chain; the targeted arm's held-out score climbs from $0.10$
to $0.89$ purely as its coverage grows, while uniform stays flat at $0.53$.

This \emph{strengthens} the coverage account. Where targeting eventually wins
($V_d\!\to\!1$) it does so because the streaming verifier has already supplied the per-turn coverage signal:
targeting is then a conduit for coverage the verifier provides, not a competing principle. The whole surface is one
coverage mechanism, with $V_d$ setting how that coverage is best delivered. It also localizes the cause of
$\kk=1$: even the state-dependent chain crosses over under a streaming verifier, so $\kk=1$ in real $\tau^2$ is a
property of the \emph{batched} verifier, not of multi-turn structure.

\paragraph{Probing the high-$V_d$ end on the real model.} On these tasks the only way to raise $V_d$ at fixed $\C$
is to inject oracle knowledge---either exposing gold read-correctness to credit, or injecting the prerequisite
target ids that make reads structurally checkable. Neither construction reproduces the toy crossover on Qwen3-8B:
even with $V_d$ raised by oracle labels, concentrated credit stays at or below uniform, so at the natural
low-$V_d$ operating point the uniform~$\ge$~concentration ordering holds \emph{a fortiori}. That \emph{both}
constructions must inject oracle knowledge to move $V_d$ at all is
itself the point---the cost the write-fraction ceiling (\S\ref{sec:writefrac}) identifies as structural.

\subsection{The write-fraction ceiling}
\label{sec:writefrac}
The need to inject oracle knowledge to move $V_d$ is not specific to $\tau^2$: it is a property of
\emph{state-based} verification. A verifier that
scores a trajectory by its effect on world state can attribute correctness only to steps that \emph{change} state; a
read or query returns information but leaves no state delta, so whether it advanced the (a-priori-unknown) goal is
invisible to the verifier unless it is supplied an oracle label or a learned judge. The independently checkable,
non-oracle fraction of a task's causal chain is therefore upper-bounded by its \emph{write-fraction}, with read and
exploration steps a structural dead zone:
\begin{equation}
V_d \;\le\; \text{write-fraction}\qquad\text{(state-transition verifier, absent oracle or learned per-step labels).}
\label{eq:writefrac}
\end{equation}
The census of Table~\ref{tab:census} shows the ceiling is binding in practice: no widely used multi-turn tool-agent
benchmark exposes a naturally high $V_d$. The measured maximum is BFCL~V3, whose per-turn checker fires only on
state-changing turns and reaches $V_d\!\approx\!0.4$; the terminal-verifier majority sits at
$V_d\!\approx\!0.1$--$0.2$. High-$V_d$ per-step signal appears only outside multi-turn tool use---in single-turn
reasoning with a \emph{learned} process verifier, or in simulator domains with an authored per-step score---that is,
where the per-step label is \emph{supplied} rather than structurally free. Two consequences follow. First, the
coverage regime ($\C\!\gg\!\kk$) is the structurally common case for tool agents, not an artifact of our task
selection, so uniform redistribution is their forced zero-information default. Second, the learned and oracle
process-reward methods that beat outcome-only training on such tasks are best read not as free streaming verifiers
but as \emph{paying} this ceiling's price---supplying the missing per-step signal (\S\ref{sec:related}).

\subsection{Prescription}
For terminal-state verifiers the observable credit signal structurally collapses to a final-write spike
($\kk=1$), while success depends on a $5$--$8$ step prerequisite chain. In this $\C\gg\kk$ regime uniform
redistribution is the zero-information coverage strategy that no fixed-budget concentration beats: a strong
default to reach for before adding learned per-turn signal, not a baseline to improve on. Learned calibration
that recovers the causal chain is the complementary lever (\S\ref{sec:related}).
The verifier-density boundary (\S\ref{sec:vd}) makes the condition concrete: concentration is worth its coverage
risk only once the verifier exposes most of the causal chain ($V_d$ near $1$), which terminal-state verifiers
structurally do not---so for the dominant paradigm, uniform coverage remains the default.

\section{Validation and boundary conditions}
\label{sec:boundary}

The mechanism predicts where the effect should \emph{hold}, where it should \emph{vanish}, and how it should
\emph{scale}; we confirm all three---and that the effect reproduces on a second benchmark, with structural
ingredients invariant across model families. One
mechanism organizes the positives, the scaling, and the nulls into a single regime map (Table~\ref{tab:regime}).

\begin{table}[h]
\centering\small
\begin{tabular}{ll >{\raggedright\arraybackslash}p{0.33\textwidth}}
\toprule
Prediction & Regime & Observed (anchors) \\
\midrule
uniform $>$ binary & above floor, $\C\gg\kk$ & $+0.079$ ($4/5$; one-sided Wilcoxon $p{=}0.003$, $95$ pairs; 14B); $+0.048$, $3/3$ (8B) \\
per-turn $<$ uniform & above floor, $\C\gg\kk$ & $5/5$ (14B); $2/3$ (8B) \\
shuffled $\approx$ per-turn & above floor & CI $[-0.07,+0.10]$ ($\tau^2$); $\Delta{=}+0.002$ (BFCL) \\
$\kk{=}1$, $\C{=}5$--$8$ & terminal verifier & median $\kk{=}1$, $\C{=}5$--$8$ (Qwen3 + Llama-3.1) \\
null (no usable gradient) & below floor / thin signal & null (4B, Llama-3.1, telecom) \\
concentration $<$ uniform, $V_d\!<\!V_d^\ast$ & 2nd benchmark, $V_d\!\approx\!0.4$ & per-turn $-0.036$ ($6/8$), shuffled $-0.034$ ($8/8$), $8$ seeds (BFCL) \\
\quad same effect, cross-family & non-Qwen, above floor & $-0.048$, $t{=}-6.54$, CI $[-0.063,-0.033]$, $32$ seeds (ToolACE) \\
coverage widens $\Rightarrow$ deficit shrinks & matched budget, above floor & $-0.048\,(\kk{=}1)\rightarrow-0.017\,(\kk{\approx}3)\rightarrow+0.010$ (all calls) vs.\ full coverage; both steps $p{<}0.01$ ($20$ seeds, ToolACE) \\
\bottomrule
\end{tabular}
\caption{One coverage mechanism predicts the positives, the scaling, \emph{and} the nulls. The structural
ingredients ($\kk{=}1$, $\C{=}5$--$8$) and the below-floor null both reproduce across model families; the
credit-geometry effect is measured within the above-floor Qwen3 family \emph{and} reproduced across family on
ToolACE-2-8B (BFCL~V3, last row). The BFCL rows are within-run paired same-batch comparisons
(\S\ref{sec:boundary}); the ToolACE row pools the pre-registered $8$-seed main set, $4$-seed replication, and
independent $20$-seed replication ($32$ seeds); the breadth row is the pre-registered matched-budget sweep on the same
frozen rollouts ($20$ seeds; Appendix~\ref{app:breadth}).
Seed counts are fractions of seeds satisfying the prediction.}
\label{tab:regime}
\end{table}

\begin{figure}[t]
\centering
\includegraphics[width=0.92\textwidth]{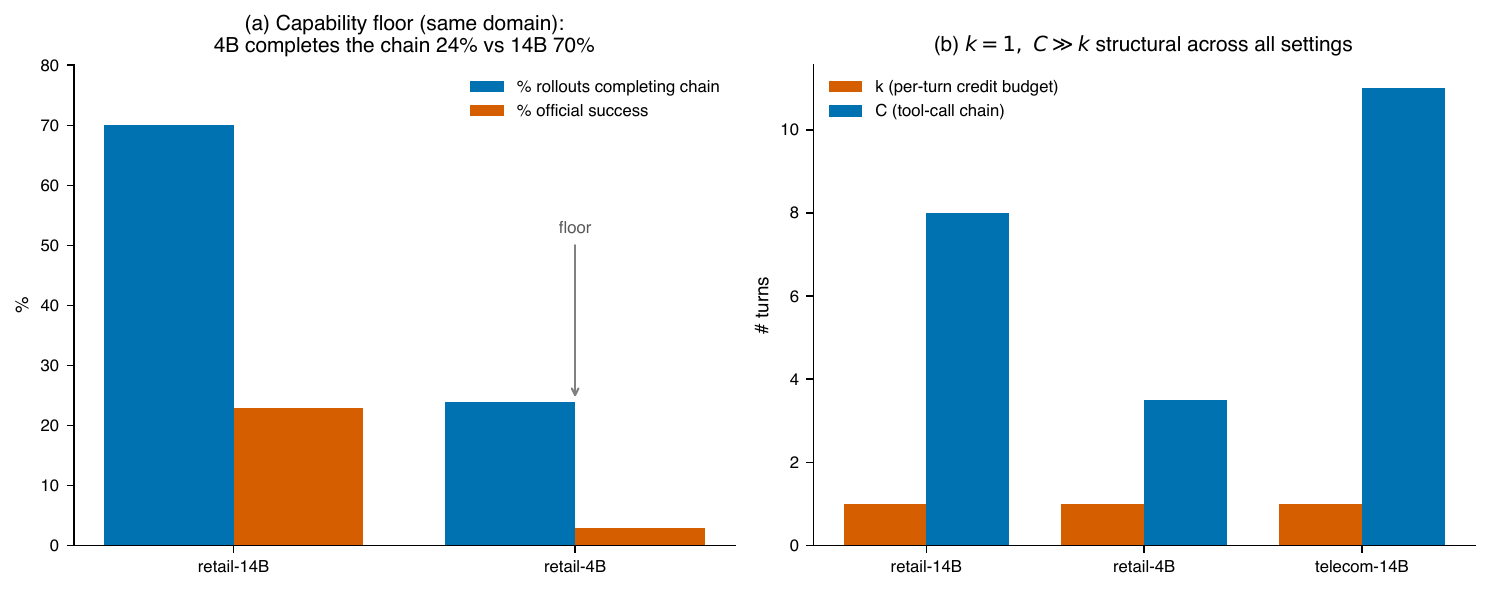}
\caption{\textbf{Boundary conditions.} (a) Capability floor: 14B completes the prerequisite chain far more often
than 4B. (b) The $\kk=1$, $\C\gg\kk$ structure holds across retail-14B, retail-4B, and telecom-14B.}
\label{fig:settings}
\end{figure}

\paragraph{Capability floor.} At 4B the comparison is null---but the diagnostic shows why: the 4B agent completes
the prerequisite chain in only $24\%$ of rollouts (vs.\ $\sim 70\%$ at 14B). With too few completed chains, no
credit geometry has anything to cover. The floor is a \emph{completion} limit, not a shortened chain: the
task-required chain is unchanged, and Fig.~\ref{fig:settings}b plots the chain each agent actually traverses.

\paragraph{Scale reproduction (within Qwen3).} Above the floor the effect reproduces and scales with capability:
dense $>$ binary is $+0.048$ at 8B ($3/3$) and $+0.079$ at 14B; per-turn $-$ uniform is $-0.050$ at 8B ($2/3$)
and $-0.053$ at 14B. Three scale anchors (4B floor / 8B / 14B) rule out a 14B-specific artifact
(4B and 14B plotted in Fig.~\ref{fig:settings}; 8B in text).

\paragraph{Cross-family structural invariance (and a predicted null).} The regime's structural ingredients are task-defined, not
Qwen-specific. On Llama-3.1-8B---an independently pretrained family---the same retail tasks traverse a causal
chain of the same length ($\C=5$--$8$; median $6$ distinct tools, $7$ on the successful rollout), and the progress
signal reaching the terminal write again collapses to $\kk=1$. Llama is genuinely doing tool use ($63\%$ of its
turns issue tool calls), so this is a structural measurement, not a parsing artifact. The credit-geometry effect
itself is \emph{not identifiable} on Llama: it sits below this benchmark's capability floor ($2\%$ base
success, comparable to the below-floor Qwen3-4B; Fig.~\ref{fig:settings}a)---consistent with public
$\tau$-bench leaderboard results \citep{taubench}, where non-Qwen agents of this class score in
the single digits---so its training groups are near-uniformly zero-variance. Its failures trace to weak grounding:
only $50\%$ of its terminal-write arguments come from its precursor reads (vs.\ $100\%$ for the above-floor Qwen
agent), i.e.\ it traverses the chain but does not reliably consume it. The regime ($\kk=1$, $\C\gg\kk$) is thus
cross-family. The absent effect is itself a prediction of the coverage account, not a failed replication: below
the floor, too few rollouts complete the chain for any credit geometry to have a gradient to exploit, so the
mechanism correctly anticipates \emph{both} where the effect appears (above-floor Qwen3) and where it vanishes
(below-floor Llama-3.1 and 4B). Scale alone does not fix this grounding deficit: Llama-3.3-70B scores $0/32$ on the
same tasks with the same traverses-but-does-not-consume signature, despite well-formed tool calls on $81\%$ of its
rollouts. The above-floor cross-family confirmation of the credit-geometry \emph{effect} comes on the second
benchmark: on BFCL~V3, where a tool-specialized Llama-3.1 agent clears the floor, concentration again underperforms
uniform (below). (On $\tau^2$-bench itself, no open non-Qwen family we tested clears the floor.)

\paragraph{Resolution gate.} In the telecom domain, where the dense signal's intermediate material is too thin to
separate arms, the comparison is null. This is the second axis---signal density---gating the \emph{visibility} of
the coverage mechanism. The mechanism itself ($\kk=1$, $\C\gg\kk$) holds structurally across all three settings:
retail-14B, retail-4B, and telecom-14B.

\begin{figure}[t]
\centering
\includegraphics[width=0.7\textwidth]{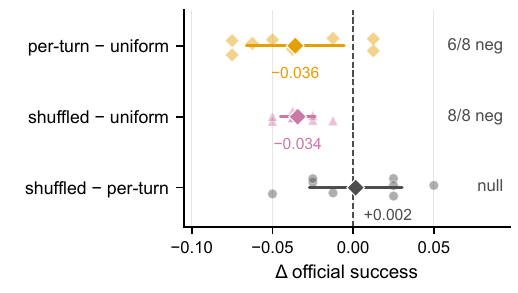}
\caption{\textbf{Second-benchmark reproduction on BFCL~V3} ($8$ paired seeds; faint markers are per-seed
differences, diamonds means with paired $95\%$ CIs). Per-turn (checker-visible) and shuffled are both harmful relative
to uniform, while the direct shuffled$-$per-turn contrast is null, reproducing the targeting null---concentration is
the harm, targeting second-order, exactly as on $\tau^2$. All arms share each seed's training rollouts and a single
fresh evaluation batch (Appendix~\ref{app:evalvar}).}
\label{fig:bfcl}
\end{figure}

\paragraph{Second-benchmark reproduction (BFCL).} Finally, we test the coverage prediction outside
$\tau^2$-bench entirely. BFCL~V3 multi-turn \citep{bfcl} has a state-based checker that verifies the backend
state \emph{after every turn}---a natively per-turn, non-oracle verifier. Measuring its structure shows that even
this checker exposes only the state-\emph{changing} steps, a median of $2$ of a $5$--$6$-step tool chain
($V_d\!\approx\!0.4$), well below the synthetic crossover $V_d^\ast\!\approx\!0.8$ (\S\ref{sec:vd})---so the coverage account predicts uniform should win here
too, and it does---and the full \S\ref{sec:targeting} targeting-falsification structure reproduces with it. In a
same-batch four-arm comparison ($8$ seeds), restricting the shared-rollout advantage to the checker-visible steps
(per-turn concentration) \emph{underperforms} spreading it uniformly across all turns (the same uniform geometry
as on $\tau^2$) on held-out official success: $\Delta(\text{per-turn}-\text{uniform})=-0.036$ (paired $t=-2.80$,
$p=0.027$; $6/8$ seeds negative; a continuous per-step support metric trends the same way). Redistributing the
\emph{same} concentrated credit onto \emph{random} turns is equally harmful---$\Delta(\text{shuffled}-\text{uniform})=-0.034$ ($8/8$ seeds negative, $t=-7.5$, $p<0.001$)---and shuffled minus per-turn is $+0.002$, with its
$95\%$ interval contained within $\pm0.031$, below the concentration deficit itself: the same targeting null as on
$\tau^2$, now on a second benchmark with a natively per-turn verifier (Fig.~\ref{fig:bfcl}). The comparison is within-run paired: all arms share each seed's training rollouts
and the same fresh evaluation batch, so between-batch evaluation drift cancels in the difference
(Appendix~\ref{app:evalvar}). The reproduction also carries a descriptive fact that does not depend
on the RL outcome: \emph{even natively streaming state verification leaves the majority of the causal chain
unobserved}, placing a second benchmark in the coverage regime---and there, as on $\tau^2$, uniform coverage
wins.

\paragraph{Cross-family reproduction of the effect (BFCL).} Having reproduced the effect on a second benchmark
within Qwen3, we ask whether it is confined to Qwen3 at all. We repeat the shared-rollout uniform-vs-concentrated
comparison on \textbf{ToolACE-2-8B} \citep{toolace}, a tool-specialized model in the independently pretrained
Llama-3.1 family ($38.4\%$ BFCL~V3 multi-turn official), which satisfies the same pre-registered eligibility
criteria as the Qwen3 runs: valid tool-call parsing, training above the variance floor, and update dynamics
matched to Qwen3.\footnote{Concretely, parser validity $0.556$, non-degenerate training success $0.281$, and
pre-registered update-dynamics parity $0.996$; the training subset was chosen by the \emph{same} model-specific
variance screen used for Qwen3, with the held-out $20$ tasks held identical (train-task overlap Jaccard $0.167$),
so the comparison is procedure-matched rather than task-matched. Non-tool turns, which ToolACE's chat template
serializes without an empty tool-call field, were normalized identically across both arms.} Concentrating credit on the checker-visible steps again underperforms
uniform spreading. The pre-registered $8$-seed main set is negative on $7/8$ seeds
($\Delta(\text{concentrated}-\text{uniform})=-0.034$, $t=-2.20$)---a magnitude matching the within-Qwen3 BFCL
effect, though its $95\%$ CI narrowly includes zero ($[-0.071,+0.003]$). Following the pre-registered analysis
plan we then ran a $4$-seed replication, which returned $4/4$ negative ($\Delta=-0.044$, $t=-2.65$), and a further pre-registered $20$-seed replication that is significant on its own ($\Delta=-0.054$, $t=-5.73$, $95\%$ CI $[-0.073,-0.034]$; $18/20$ negative). The pooled
$32$-seed estimate is $\Delta=-0.048$ ($t=-6.54$, $95\%$ CI $[-0.063,-0.033]$; $29/32$ negative). We report the
main set, each replication, and the pooled estimate separately, as pre-registered. An above-floor non-Qwen agent
thus shows the same concentration-underperforms-uniform result: the credit-geometry effect is not confined to a
single model family.

\begin{figure}[t]
\centering
\includegraphics[width=0.7\textwidth]{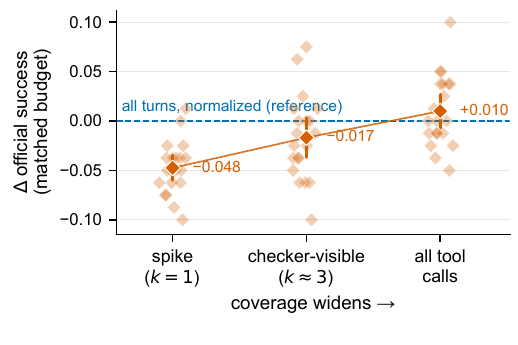}
\caption{\textbf{Matched-budget coverage dose--response} on the frozen ToolACE rollouts ($20$ seeds; official
success). Coverage widens left to right at fixed total credit $\sum_t|a_t|=|A_i|$; faint markers are per-seed
differences against the all-turns-normalized reference (dashed line), diamonds means with paired $95\%$ CIs. The
deficit shrinks monotonically and vanishes at full chain coverage; adjacent steps are significant at $p<0.01$
(Appendix~\ref{app:breadth}).}
\label{fig:breadth}
\end{figure}

\paragraph{Breadth dose--response.} The coverage account makes a finer prediction than a two-point contrast: at a
fixed total credit budget, performance should rise monotonically with the fraction of the chain covered. A
pre-registered five-arm sweep on the same frozen ToolACE rollouts confirms it (Fig.~\ref{fig:breadth}; $20$ seeds;
official success; all arms share each seed's training batch and are evaluated in one common batch per seed; full
table and protocol in Appendix~\ref{app:breadth}). Holding $\sum_t |a_t|=|A_i|$ fixed, concentrating the advantage on
a single turn costs $-0.048$ relative to spreading it over all turns ($a_t=A_i/n$; $t=-7.84$; $18/20$ negative);
widening to the checker-visible steps recovers $+0.031$ ($t=3.12$), and widening further to every tool call recovers
another $+0.027$ ($t=3.05$), reaching parity with full coverage ($+0.010$, $95\%$ CI $[-0.007,+0.027]$). The
targeting null reproduces inside the sweep: shuffling the checker-visible credit onto random turns shifts it by a
statistically indistinguishable $+0.014$ ($90\%$ CI within $\pm0.037$) while both arms stay below uniform. Doubling
the concentrated arm's learning rate---driving its post-update KL two orders of magnitude above uniform's---still
leaves it $-0.024$ below uniform ($t=-2.70$). Coverage and scale decompose cleanly: breadth accounts for the ladder
above, and the standard un-normalized allocation contributes a further $+0.033$ over the all-turns-normalized variant
($t=3.08$)---full chain coverage is necessary, and the standard terminal advantage completes the gain.

\paragraph{Reward-to-go restores prefix coverage.} The account also makes a prediction about the classical remedy
for delayed credit: reward-to-go redistribution should succeed \emph{because it restores prefix coverage}, not because
it targets better. With a purely terminal reward, per-turn returns-to-go are constant across the trajectory---the
classical estimator already \emph{is} the uniform allocation in this regime; concentration arises only when an
intermediate progress signal invites it. A pre-registered reward-to-go arm built from the intermediate progress
signal, on the BFCL Qwen3-8B rollouts ($5$ seeds, same frozen-batch, common-batch protocol), confirms the prediction:
reward-to-go lands at parity with full coverage ($+0.025$ vs.\ the normalized all-turns arm, $95\%$ CI
$[-0.019,+0.069]$) and above checker-visible concentration ($+0.045$, CI $[+0.003,+0.087]$;
Appendix~\ref{app:breadth}).

\paragraph{The coverage regime is the common case.} BFCL is not an outlier. Table~\ref{tab:census} surveys the verifier structures of
widely used agentic-RL benchmarks: measured $V_d$ for the two we measure directly (this work), and the
verifier's structural type elsewhere. Terminal-state verification predominates---for such verifiers, the
collapse of observable per-turn progress to the final step ($\kk{=}1$) is an analytical property of the reward
code, not an empirical parameter---and the two directly measured points, $\tau^2$'s terminal DB check
($V_d\!\approx\!0.15$) and BFCL~V3's natively per-turn state checker ($V_d\!\approx\!0.4$), both fall well
below the crossover $V_d^\ast\!\approx\!0.8$. Even ToolSandbox, the intermediate-milestone exception, exposes
only part of the chain.

\begin{table}[h]
\centering\footnotesize
\setlength{\tabcolsep}{4pt}
\begin{tabular}{llll}
\toprule
Benchmark & Verifier & Checked & $V_d$ \\
\midrule
$\tau^2$-bench \citep{tau2} & final DB state & terminal & \textbf{0.12--0.20 (measured)} \\
BFCL V3 multi-turn \citep{bfcl} & backend state per turn (writes only) & per-turn & \textbf{$\approx$0.4 (measured)} \\
SWE-bench \citep{swebench} & post-patch test suite & terminal & \textbf{$\kk{=}1$ (from scoring code)} \\
WebArena \citep{webarena} & final page/URL state (state-scored tasks) & terminal & \textbf{$\kk{=}1$ (from scoring code)} \\
OSWorld \citep{osworld} & final system state & terminal & \textbf{$\kk{=}1$ (from scoring code)} \\
AppWorld \citep{appworld} & final-state unit tests & terminal & low (structural) \\
ToolSandbox \citep{toolsandbox} & trajectory milestones & intermediate & partial \\
ScienceWorld \citep{scienceworld} & simulator score per step & per-step & high (non-tool domain) \\
\bottomrule
\end{tabular}
\caption{Verifier structures across agentic benchmarks. ``low (structural)'' denotes terminal-state verification,
which collapses the observable per-turn signal to $\kk{=}1$ (\S\ref{sec:diag}); measured values are from this
work. For SWE-bench, WebArena (state-scored task types), and OSWorld, $\kk{=}1$---and hence
$V_d=1/\C\le$ write-fraction (Eq.~\ref{eq:writefrac})---follows analytically from the official scoring harnesses,
which invoke the task verifier once on the terminal state (pinned commits: SWE-bench \texttt{f7bbbb2},
\texttt{run\_evaluation.py}; WebArena \texttt{dce0468}, \texttt{evaluators.py}; OSWorld \texttt{7a17d3a},
\texttt{desktop\_env.py}). The coverage regime is the common case.}
\label{tab:census}
\end{table}

\section{Related work}
\label{sec:related}
\textbf{Adding intermediate signal.} Turn-level reward methods, turn-level critics, and learned per-step
calibration \citep{mtgrpo,irc,archer,sweetrl,vpr} add \emph{new} intermediate supervision---a \emph{learned}
per-turn signal, in the tradition of process supervision \citep{lightman}. Our harmful
arm is a \emph{naive structural} concentration (progress-derived $\Delta S$) with zero new information, and our
coverage account predicts exactly when such methods
help: a learned signal wins to the extent it restores coverage of the causal chain that the structural signal
collapses ($\kk{=}1$). In the language of \S\ref{sec:vd}, such methods raise an \emph{effective}, data-driven
$V_d$---a learned model recovering per-turn signal, subject to its own bias and the localization risk below---which
is distinct from the structural $V_d$ swept in \S\ref{sec:vd}. Their gains over outcome-only baselines are thus
\emph{compatible with} the high-$V_d$ prediction. We characterize the coverage gap such methods must close, and the $\C\gg\kk$
regime in which zero-information uniform spreading is already a strong default. We further note that a learned signal that
\emph{localizes} credit onto a few turns inherits the same coverage risk our structural concentration arms expose:
localization is unreliable exactly in the $\C\gg\kk$ regime that terminal-verifier agents occupy. Structural
step-level schemes \citep{gigpo} construct fine-grained per-step advantages from repeated states without new
supervision; our account predicts that their benefit tracks the chain coverage the step-level signal restores, and the
matched-concentration shuffled control is the guard that separates coverage from targeting in any fixed-budget
redistribution. Classical credit assignment---potential-based shaping \citep{ng1999}, return decomposition \citep{rudder}, and hindsight credit assignment \citep{hca}---reshapes where credit lands; at $\C\gg\kk$, however, coverage breadth dominates that location. \textbf{Reward
richness.} Tool-use reward shaping and the sparse-to-dense result \citep{toolrl,hybridrl} establish that richer
feedback can help; our contribution is orthogonal, isolating the geometry of how the same terminal advantage is
distributed. \textbf{Token-level uniformity.}
Sequence-level control of token-level ratio variance within one response \citep{gspo} is a mechanistic precedent at a
different granularity; our contribution is the multi-turn rollout coverage account and its phase boundary.
\textbf{Coverage over credit routing.} A recent report reaches a parallel conclusion at the module level
\citep{ge2026coverage}: routing a fixed zeroth-order perturbation budget across the modules of a frozen tool-using
LLM agent by trajectory-level failure credit does not improve on-pool sample efficiency over uniform allocation, the
loss under credit-routed schedules scales with how often the bottleneck module is starved of updates, and a
pre-registered credit-free coverage floor removes the detected harm. That setting routes a perturbation budget across
\emph{parameters} under gradient-free optimization rather than a terminal advantage across \emph{turns} under policy
gradients; the shared conclusion---uniform coverage is the default that credit-routed concentration must beat---is
complementary to ours.

\section{Limitations}
\label{sec:limits}
The mechanism is characterized for terminal-verifier tool agents in the $\C\gg\kk$ regime---the common case for
DB/state-verified agents, above the capability floor. The structural ingredients ($\kk=1$ with $\C\gg\kk$;
$\C=5$--$8$ on retail) are
measured \emph{across two model families} (Qwen3 and Llama-3.1) and three $\tau^2$ task settings (retail-14B,
retail-4B, telecom-14B), and the phase boundary is
established in a domain- and algorithm-agnostic toy, so the account is not retail- or Qwen-specific. The
credit-geometry \emph{effect} is confirmed across model families on BFCL~V3---a tool-specialized Llama-3.1 agent
(ToolACE-2-8B) reproduces concentration $<$ uniform above the floor and passes the matched-budget breadth sweep
(\S\ref{sec:boundary})---but on $\tau^2$-bench itself it is observed only within Qwen3, the only open family of this
scale class that cleared that benchmark's floor in our tests.

Our real-model comparisons isolate the first policy update; longer-horizon training dynamics are outside the
scope of this study. The
toy---where concentration wins at $\C\le\kk$ \emph{despite} higher per-update variance---rules out per-token
gradient variance (which uniform spreading also minimizes) as the sole explanation, and the matched-budget sweep
separates coverage from update scale on the real model (\S\ref{sec:boundary}). The Qwen3-8B battery
(\S\ref{sec:targeting}) is reported directionally, its arms having been trained in separate runs
(Appendix~\ref{app:evalvar}).

This also motivates a
\emph{methodological} recommendation: because changing credit geometry moves
concentration and optimization dynamics together, any claim that a learned or oracle per-turn signal helps by
better \emph{localization} (rather than by broadening coverage or changing effective step size) should be tested
against a matched-concentration shuffled control; absent that control, reported per-turn gains are confounded.

\appendix
\section{The per-turn deficit holds across the trainable dose range}
\label{app:dose}
The geometry result (per-turn $<$ uniform) could in principle be specific to the single optimizer dose used in
the main experiments, or to a difference in effective update magnitude between geometries. It is not. We swept the update strength across the full range in which the policy remains
trainable (Table~\ref{tab:dose}). At the gentle main-experiment dose the per-turn arm trails uniform by $0.053$
(5 seeds). Pushing the effective dose $40\times$ (single-epoch, so the importance ratio stays near one and training
cannot diverge in-step) over-steps the policy---both arms degrade---and the per-turn deficit \emph{widens} to
$0.20$ (paired bootstrap CI $[-0.33,-0.02]$ on the seed that does not fully collapse). Raising the learning rate
under two-epoch updates instead diverges (the asymmetric treatment of negative-advantage turns across epochs blows
up the off-policy ratio), bounding high-dose stability under multi-epoch updates. Across every trainable dose,
per-turn $\le$ uniform.

\begin{table}[h]
\centering\small
\begin{tabular}{lcccc}
\toprule
Dose regime & rel.\ LR & stable? & $\Delta$(per-turn $-$ uniform) & seeds \\
\midrule
Gentle (main) & $1\times$ & yes & $-0.053$ & 5 \\
Aggressive (single-epoch) & $40\times$ & trains, policy over-steps & $-0.20$ (CI $[-0.33,-0.02]$) & 2 \\
High-LR, two-epoch & $10$--$40\times$ & diverges & --- (untrainable) & 2 \\
\bottomrule
\end{tabular}
\caption{Per-turn concentration underperforms uniform across the entire trainable dose range.
$\Delta$ on held-out assert-support; negative $=$ per-turn worse.}
\label{tab:dose}
\end{table}

\section{Between-batch evaluation variability, and why all comparisons are within-run paired}
\label{app:evalvar}
Our held-out evaluation samples $4$ rollouts per task over $19$ tasks. To quantify the run-to-run variability of
this estimator we re-evaluated the \emph{same} four trained uniform adapters on the \emph{same} task set in a
fresh evaluation batch under an identical configuration (same split, sampling, serving stack, and user
simulator). Per-seed assert-support moved by $+0.061$, $-0.042$, $+0.112$, and $+0.159$ between batches, and a third
re-evaluation of the same adapters spans a range of about $0.07$. The same instability recurs on BFCL:
re-evaluating the same trained adapters in a fresh batch shifts official success by a mean of $-0.009$ over a
range of $[-0.05,+0.04]$ across seeds---itself the size of the credit-geometry effect we measure there, which is
exactly why the BFCL arms (including the shuffled control) are compared within-run paired. These
measurements motivate our reporting rule: confirmatory numeric arm comparisons use a \textbf{within-run paired}
design---competing credit geometries share the same training rollouts and are evaluated on identical task
instances in the same batch, so batch-level environment and simulator noise affects both arms equally and
cancels in the paired difference (this holds for the main $\tau^2$ arms, the $V_d$ arms, and the BFCL
reproduction). Inference likewise respects the seed hierarchy: for uniform $-$ binary, a paired hierarchical
bootstrap that first resamples the five seeds and then the $19$ paired tasks within each selected seed gives a mean
difference of $+0.0789$ with percentile $95\%$ CI $[+0.0085,+0.1492]$ ($20{,}000$ replicates, fixed random seed).
Auxiliary arms trained and evaluated in separate runs (the gold/random/write/read battery of
\S\ref{sec:targeting}) are reported directionally, not as effect-fraction estimates. A matched-batch paired
design is the inexpensive guard we recommend for per-turn credit-assignment comparisons at small evaluation
budgets.

\section{Synthetic environment specification}
\label{app:toyspec}
The controlled environment of \S\ref{sec:toy} and \S\ref{sec:vd} is a $T$-turn episodic task with known causal
structure, small enough to sweep exhaustively. \textbf{Task.} Each episode has $T{=}12$ turns ($T{=}24, 48$ in
the horizon sweep); turn $t$ presents a fixed random unit-norm feature $\phi_t\in\mathbb{R}^8$ and the policy
picks one of $K$ actions ($K{=}5$; $K{=}10$ in the $V_d$ sweep, which keeps the uniform arm off its performance
ceiling so the crossover location is visible). A hidden subset of $\C$ turns is causally necessary, each with
one designated correct action. Under the \emph{independent} structure the return is the fraction of causal
turns answered correctly; under the \emph{chain} structure a causal turn counts only if all earlier causal
turns are also correct, mirroring the prerequisite reads of \S\ref{sec:diag}; a \emph{graded} variant replaces
the binary causal set with per-turn relevance weights. \textbf{Policy and update.} The policy is a two-layer
MLP (hidden width $16$, parameters shared across turns) mapping $\phi_t$ to a softmax over actions. Each update
samples a group of $G{=}24$ rollouts, computes the group-normalized terminal advantage $A_i$ (GRPO),
distributes it across turns according to the arm, and applies a policy-gradient step; training runs $400$ steps
at learning rate $0.6$ under a global gradient-norm clip of $1.0$, which matches the effective step size across
arms and algorithms. The algorithm-agnostic check replaces the group normalization with a batch-mean baseline
and with plain REINFORCE at the same learning rate and clip. \textbf{Arms.} Uniform spreads $A_i/T$ over all
turns (the budget-normalized counterpart of the real-model sweep's all-turns arm). Concentrated places the credit on a budget of $\kk$ turns ($\kk{=}3$ in the $\C$-sweep over
$\C\in\{1,2,3,4,6,8,10,12\}$); in the $V_d$ sweep the targeted arm receives an independent group-normalized per-turn
signal on the verifier-exposed causal turns plus the terminal residual on the final causal turn. Every arm is
rescaled per rollout so that its total absolute credit equals $|A_i|$: arms differ only in credit shape, never
in magnitude. \textbf{Scale.} The $\C$-sweep uses $6$--$12$ paired seeds per point; the $V_d$ sweep uses $32$
paired seeds; $\kk_{\text{eff}}$ is the participation ratio of the mean per-turn absolute credit.
Re-running the $V_d$ sweep at hidden widths $8$ and $32$ and action counts $K{=}5$ and $K{=}20$ leaves the crossover
within $[0.79,0.88]$ (baseline $0.81$).

\section{Real-model training configuration}
\label{app:config}
All real-model comparisons use the same GRPO/LoRA stack (harness names in the footnote of \S\ref{sec:setup});
per-benchmark settings are below. Within a benchmark, arms differ \emph{only} in the reward argument.
The $\tau^2$ experiments run on single data-center GPUs of the NVIDIA H100/H20 class ($94$--$96$\,GB); the BFCL
and ToolACE experiments, including the breadth and reward-to-go sweeps, run on single $96$\,GB NVIDIA RTX PRO 6000
workstation GPUs. vLLM serves rollout collection and evaluation and PyTorch/PEFT performs LoRA training on Linux
(libraries listed in the released harness); each (arm, seed) chain fits on one GPU.
PPO-ratio clipping is essentially inactive throughout: the clipped-token fraction stays below $1\%$ in every $\tau^2$
and BFCL arm and below $2.5\%$ on ToolACE, with no arm separation. At the main dose the two $\tau^2$ geometries also
displace the policy equally: per-seed post-update KL from the base policy lies in the same
$5\times10^{-4}$--$1.1\times10^{-3}$ band for uniform and per-turn. Group-level trainability separates the reward
types: under binary, $0.50$--$0.75$ of $\tau^2$ training groups have nonzero advantage variance (mean $0.65$), versus
$0.75$--$1.00$ (mean $0.90$) under the dense arms---quantifying the non-degenerate group signal of
\S\ref{sec:density}.

\begin{table}[h]
\centering\small
\begin{tabular}{lll}
\toprule
 & $\tau^2$-bench & BFCL~V3 multi-turn \\
\midrule
Optimizer & GRPO (clip $0.2$) & GRPO (clip $0.2$) \\
Gradient-norm clip & $1.0$ & $1.0$ \\
LoRA rank & $16$ & $16$ \\
Learning rate & $5\times10^{-5}$ & $5\times10^{-5}$ \\
KL coefficient & $10^{-3}$ & $6\times10^{-4}$ \\
Epochs per update & $2$ & $2$ \\
Training batch & $8$ tasks $\times$ group $6$ & $8$ tasks $\times$ group $4$ \\
Max sequence length (train) & $8{,}000$ & $8{,}000$ \\
Held-out evaluation & $19$ gradable tasks $\times$ $4$ & $20$ tasks $\times$ $4$ \\
Sampling temperature (eval) & $0.7$ & $0.7$ \\
\bottomrule
\end{tabular}
\caption{Training and evaluation configuration for the real-model comparisons.}
\label{tab:config}
\end{table}

\section{Metric definition and evaluation protocol}
\label{app:metric}
\paragraph{Formal definition of held-out assert-support.}
For task $q$, let $d_q^0$, $d_q^\star$, and $\hat d_{saqj}$ be the flattened initial, gold-replayed, and predicted
terminal database states for seed $s$, arm $a$, and rollout $j$. Define the gold-modified existing-field set
\[
  M_q=\{k\in\operatorname{dom}(d_q^0)\cap\operatorname{dom}(d_q^\star):
  d_q^0(k)\ne d_q^\star(k)\}.
\]
The rollout score is
\[
  S_{saqj}=\frac{1}{|M_q|}\sum_{k\in M_q}
  \mathbf 1\!\left[\hat d_{saqj}(k)=d_q^\star(k)\right],
\]
where a missing predicted key is read as its initial value $d_q^0(k)$; if $M_q$ is empty the evaluator falls back to
the official binary outcome. This evaluation quantity is distinct from the training-time per-turn support, whose
target set also includes gold-created keys. For each task we average its four rollout scores, then take an unweighted
macro-average over the $19$ frozen gradable tasks; reported contrasts average within-seed differences of these
task-macro means, and no rollout or task--seed cell is treated as an independent seed.

\paragraph{Frozen gradable subset.}
The pre-results task-property rule is $|G_q|\ge2$, where $G_q$ counts all gold DB changes including created fields.
It selects exactly the held-out task IDs $1$, $9$, $11$, $22$, $30$, $49$, $56$, $73$, $74$, $75$, $81$, $85$, $86$,
$90$, $91$, $93$, $94$, $98$, and $100$. Official task success is computed over all $20$ held-out tasks;
assert-support over the $19$ gradable ones.

\paragraph{User environment and decoding.}
In the $\tau^2$ campaign the user simulator is the fixed base Qwen3-14B served from the same local vLLM endpoint as
the evaluated agent, with temperature $0.5$, at most $512$ generated tokens per user turn, and thinking disabled;
evaluation seeds also seed the serving stack. The BFCL campaign has no generative user simulator: each user turn is
the benchmark's fixed multi-turn prompt, and the evaluated Qwen3-8B agent samples at temperature $0.7$ with at most
$512$ tokens.

\section{Additional coverage and targeting diagnostics}
\label{app:diag}
\paragraph{The causal-chain length is insensitive to its definition.}
Table~\ref{tab:cdef} recomputes $\C$ under three definitions---every executed tool call, distinct executed tools,
and the data-flow ancestors of the terminal write (precursor reads plus the write)---on the retained raw-trace
archive of official-success $\tau^2$-retail rollouts. All three plug-in $V_d$ values are at most $0.167$, consistent
with the main-text diagnostic range ($\C=5$--$8$; $V_d=0.12$--$0.20$) and roughly fivefold below the synthetic
crossover.

On the same archive, $72$ of $76$ unique identifiers consumed by the terminal writes ($94.7\%$) do not appear in the
user instruction, so the write's arguments cannot be assembled from the instruction alone.

\begin{table}[h]
\centering\small
\begin{tabular}{@{}lcc@{}}
\toprule
Definition of $\C$ & Median [IQR] & $V_d$ \\
\midrule
Executed tool calls & $8\ [6,11.25]$ & $0.125$ \\
Distinct executed tools & $6\ [4.75,6.25]$ & $0.167$ \\
Provenance ancestors & $6\ [3.75,7.25]$ & $0.167$ \\
\bottomrule
\end{tabular}
\caption{Sensitivity of the causal-chain length $\C$ to its definition ($16$ official-success rollouts from the
retained raw-trace archive; $V_d$ is the plug-in ratio with measured median $\kk_{\text{eff}}=1.0$). The low-density
diagnosis does not depend on how $\C$ is counted.}
\label{tab:cdef}
\end{table}

\paragraph{The geometry deficit holds on official task success.} The main-text geometry result uses held-out
gradable assert-support; it reproduces on official task success, the deployed metric. Across the five retail-14B
seeds official success is $0.41/0.34/0.44/0.43/0.40$ for uniform and $0.36/0.34/0.31/0.33/0.36$ for per-turn
(Table~\ref{tab:official}); the paired difference per-turn $-$ uniform is $-0.063$ (paired $t=-2.80$, $4/5$ negative
with one tie), the same sign and comparable magnitude as the assert-support deficit. Concentrating the terminal
advantage on progress turns underperforms uniform on the metric practitioners optimize, not only on the proxy.

\begin{table}[h]
\centering\small
\begin{tabular}{lcccccc}
\toprule
arm & s1 & s2 & s3 & s4 & s5 & mean \\
\midrule
base (untrained) & $0.41$ & $0.34$ & $0.38$ & $0.36$ & $0.35$ & $0.367$ \\
uniform  & $0.41$ & $0.34$ & $0.44$ & $0.43$ & $0.40$ & $0.403$ \\
per-turn & $0.36$ & $0.34$ & $0.31$ & $0.33$ & $0.36$ & $0.340$ \\
\midrule
$\Delta$ (per-turn $-$ uniform) & $-0.05$ & $0.00$ & $-0.13$ & $-0.10$ & $-0.04$ & $-0.063$ \\
\bottomrule
\end{tabular}
\caption{Per-seed official task success on $\tau^2$-bench retail-14B (held-out $20$ tasks). The geometry deficit
reproduces on official success: per-turn $-$ uniform $=-0.063$ (paired $t=-2.80$, $4/5$ negative, one tie), matching
the assert-support result on the deployed metric. The untrained base is evaluated in the same batches as the trained
arms: uniform gains $+3.6$pp over the untrained policy while per-turn falls $2.7$pp below it. Per-seed entries are
rounded to two decimals; means are computed from unrounded values.}
\label{tab:official}
\end{table}

\section{The pre-registered credit-breadth sweep on ToolACE}
\label{app:breadth}
The sweep reuses the frozen rollout batches of the $20$-seed ToolACE replication (\S\ref{sec:boundary}). For each
seed, five new arms---a single-turn spike ($\kk{=}1$, placed on the turn carrying maximal checker-visible progress),
the checker-visible weights shuffled onto random turns, all executed tool calls ($\kk{\approx}5$--$6$), all turns
($a_t=A_i/n$, where $n$ is the rollout's turn count), and the checker-visible arm at doubled learning rate---are
trained from the same batch under the identical configuration (Table~\ref{tab:config}), and all arms, including the
archived uniform and checker-visible-steps ($\kk{\approx}3$) adapters, are re-evaluated in a single common batch per
seed (one evaluation session, identical task order, held-out $20$ tasks $\times$ $4$ rollouts; official success;
Table~\ref{tab:breadth}). Every budget-matched arm redistributes the same group-normalized terminal advantage with
$\sum_t |a_t| = |A_i|$, enforced to numerical precision. A data-flow oracle set in the sense of the $\tau^2$
diagnostic is not constructible on BFCL (median provenance set size $1$), so the pre-registered fallback, all
executed tool calls, serves as the widest concentrated arm.

\begin{table}[h]
\centering\small
\begin{tabular}{lcc}
\toprule
arm ($\kk_{\text{eff}}$) & $\Delta$ vs.\ uniform ($a_t{=}A_i$) & $\Delta$ vs.\ all-turns normalized \\
\midrule
spike ($1$) & $-0.081$ ($t=-7.75$; $18/20$ neg.) & $-0.048$ ($t=-7.84$; $18/20$ neg.) \\
checker-visible steps ($\approx 3$) & $-0.050$ & $-0.017$ ($t=-1.76$) \\
shuffled, matched count ($\approx 3$) & $-0.036$ ($t=-2.34$) & $-0.003$ \\
all tool calls ($\approx 5$--$6$) & $-0.023$ ($t=-2.21$) & $+0.010$ (CI $[-0.007,+0.027]$) \\
all turns, normalized ($n$) & $-0.033$ ($t=-3.08$) & $0$ \\
checker-visible at $2\times$ LR & $-0.024$ ($t=-2.70$) & $+0.009$ \\
\bottomrule
\end{tabular}
\caption{Breadth sweep on ToolACE ($20$ seeds, per-seed paired, official success). Signs follow the stated contrast
direction (named arm minus reference). Adjacent-step contrasts: spike $\to$ checker-visible $+0.031$ ($t=3.12$);
checker-visible $\to$ all tool calls $+0.027$ ($t=3.05$). The $2\times$-LR arm's post-update KL is $0.60$ versus
uniform's $0.006$; it brackets the deficit from above. Re-evaluating the archived arms in the common batch reproduces
the archived contrast ($-0.050$ vs.\ $-0.054$).}
\label{tab:breadth}
\end{table}

\paragraph{Reward-to-go on Qwen3-8B.}
A pre-registered companion experiment reuses the frozen rollout batches of the BFCL Qwen3-8B campaign ($5$ seeds):
three budget-normalized breadth arms (the single-turn spike, all tool calls, and all turns) and a reward-to-go arm
($w_t \propto \sum_{t' \ge t} \Delta S_{t'}$, the classical delayed-credit remedy) are trained per seed, and all
eight arms---including the four archived ones---are evaluated in one common batch per seed. No re-evaluated archived
contrast reverses its archived direction. Reward-to-go scores highest of all eight arms ($0.340$): $+0.025$ over the
normalized all-turns arm ($95\%$ CI $[-0.019,+0.069]$; parity with full coverage), $+0.045$ over checker-visible
concentration (CI $[+0.003,+0.087]$), and $+0.020$ over progress-targeted concentration---as the coverage account
predicts, since reward-to-go propagates the terminal signal to every turn preceding the last progress point. The
concentrated-arm deficits at this scale are directionally consistent (spike $-0.018$; $4/5$ negative) but do not
resolve the intermediate rungs at five seeds; the $20$-seed matched-budget ToolACE sweep above resolves the full
staircase.

\bibliographystyle{plainnat}
\bibliography{references}

@article{toolrl,
  title   = {{ToolRL}: Reward is All Tool Learning Needs},
  author  = {Qian, Cheng and Acikgoz, Emre Can and He, Qi and Wang, Hongru and others},
  journal = {arXiv preprint arXiv:2504.13958},
  year    = {2025},
  eprint  = {2504.13958},
  archivePrefix = {arXiv}
}

@article{hybridrl,
  title   = {Hybrid Reinforcement: When Reward Is Sparse, It's Better to Be Dense},
  author  = {Tao, Leitian and Kulikov, Ilia and Saha, Swarnadeep and Wang, Tianlu and others},
  journal = {arXiv preprint arXiv:2510.07242},
  year    = {2025},
  eprint  = {2510.07242},
  archivePrefix = {arXiv}
}

@article{irc,
  title   = {Multi-Turn Reinforcement Learning for Tool-Calling Agents with Iterative Reward Calibration},
  author  = {Modecrua, Wachiravit and Kaewtawee, Krittanon and Pachtrachai, Krittin and Kraisingkorn, Touchapon},
  journal = {arXiv preprint arXiv:2604.02869},
  year    = {2026},
  eprint  = {2604.02869},
  archivePrefix = {arXiv}
}

@article{mtgrpo,
  title   = {Reinforcing Multi-Turn Reasoning in {LLM} Agents via Turn-Level Reward Design},
  author  = {Wei, Quan and Zeng, Siliang and Li, Chenliang and Brown, William and others},
  journal = {arXiv preprint arXiv:2505.11821},
  year    = {2025},
  eprint  = {2505.11821},
  archivePrefix = {arXiv}
}

@article{gspo,
  title   = {Group Sequence Policy Optimization},
  author  = {Zheng, Chujie and Liu, Shixuan and Li, Mingze and Chen, Xiong-Hui and others},
  journal = {arXiv preprint arXiv:2507.18071},
  year    = {2025},
  eprint  = {2507.18071},
  archivePrefix = {arXiv}
}

@misc{bfcl,
  title        = {{BFCL} V3: Multi-Turn and Multi-Step Function Calling},
  author       = {Fanjia Yan and Huanzhi Mao and Charlie Cheng-Jie Ji and Tianjun Zhang and Shishir G. Patil and Ion Stoica and Joseph E. Gonzalez},
  year         = {2024},
  howpublished = {Berkeley Function-Calling Leaderboard},
  note         = {\url{https://gorilla.cs.berkeley.edu/blogs/13_bfcl_v3_multi_turn.html}}
}

@article{appworld,
  title   = {{AppWorld}: A Controllable World of Apps and People for Benchmarking Interactive Coding Agents},
  author  = {Harsh Trivedi and Tushar Khot and Mareike Hartmann and Ruskin Manku and Vinty Dong and Edward Li and Shashank Gupta and Ashish Sabharwal and Niranjan Balasubramanian},
  journal = {arXiv preprint arXiv:2407.18901},
  year    = {2024}
}

@article{scienceworld,
  title   = {{ScienceWorld}: Is your Agent Smarter than a 5th Grader?},
  author  = {Ruoyao Wang and Peter Jansen and Marc-Alexandre C{\^o}t{\'e} and Prithviraj Ammanabrolu},
  journal = {arXiv preprint arXiv:2203.07540},
  year    = {2022}
}

@article{toolsandbox,
  title   = {{ToolSandbox}: A Stateful, Conversational, Interactive Evaluation Benchmark for {LLM} Tool Use Capabilities},
  author  = {Jiarui Lu and Thomas Holleis and Yizhe Zhang and Bernhard Aumayer and Feng Nan and Felix Bai and Shuang Ma and Shen Ma and Mengyu Li and Guoli Yin and Zirui Wang and Ruoming Pang},
  journal = {arXiv preprint arXiv:2408.04682},
  year    = {2024}
}

@article{tau2,
  title   = {$\tau^2$-Bench: Evaluating Conversational Agents in a Dual-Control Environment},
  author  = {Victor Barres and Honghua Dong and Soham Ray and Xujie Si and Karthik Narasimhan},
  journal = {arXiv preprint arXiv:2506.07982},
  year    = {2025}
}

@inproceedings{toolace,
  title        = {{ToolACE}: Winning the Points of {LLM} Function Calling},
  author       = {Weiwen Liu and Xu Huang and Xingshan Zeng and Xinlong Hao and Shuai Yu and Dexun Li and others},
  booktitle    = {International Conference on Learning Representations (ICLR)},
  year         = {2025},
  note         = {\url{https://arxiv.org/abs/2409.00920}}
}

@article{archer,
  title   = {{ArCHer}: Training Language Model Agents via Hierarchical Multi-Turn {RL}},
  author  = {Zhou, Yifei and Zanette, Andrea and Pan, Jiayi and Levine, Sergey and Kumar, Aviral},
  journal = {arXiv preprint arXiv:2402.19446},
  year    = {2024}
}

@article{sweetrl,
  title   = {{SWEET-RL}: Training Multi-Turn {LLM} Agents on Collaborative Reasoning Tasks},
  author  = {Zhou, Yifei and Jiang, Song and Tian, Yuandong and Weston, Jason and others},
  journal = {arXiv preprint arXiv:2503.15478},
  year    = {2025}
}

@article{gigpo,
  title   = {Group-in-Group Policy Optimization for {LLM} Agent Training},
  author  = {Feng, Lang and Xue, Zhenghai and Liu, Tingcong and An, Bo},
  journal = {arXiv preprint arXiv:2505.10978},
  year    = {2025}
}

@article{lightman,
  title   = {Let's Verify Step by Step},
  author  = {Lightman, Hunter and Kosaraju, Vineet and Burda, Yura and Edwards, Harri and others},
  journal = {arXiv preprint arXiv:2305.20050},
  year    = {2023}
}

@inproceedings{proxmo,
  title     = {Proximity-Based Multi-Turn Optimization: Practical Credit Assignment for {LLM} Agent Training},
  author    = {Fang, Yangyi and Lin, Jiaye and Fu, Xiaoliang and Qin, Cong and Shi, Haolin and others},
  booktitle = {Proceedings of the 64th Annual Meeting of the Association for Computational Linguistics (Industry Track)},
  year      = {2026}
}

@article{swebench,
  title   = {{SWE}-bench: Can Language Models Resolve Real-World {GitHub} Issues?},
  author  = {Jimenez, Carlos E. and Yang, John and Wettig, Alexander and Yao, Shunyu and others},
  journal = {arXiv preprint arXiv:2310.06770},
  year    = {2023}
}

@article{webarena,
  title   = {{WebArena}: A Realistic Web Environment for Building Autonomous Agents},
  author  = {Zhou, Shuyan and Xu, Frank F. and Zhu, Hao and Zhou, Xuhui and others},
  journal = {arXiv preprint arXiv:2307.13854},
  year    = {2023}
}

@article{osworld,
  title   = {{OSWorld}: Benchmarking Multimodal Agents for Open-Ended Tasks in Real Computer Environments},
  author  = {Xie, Tianbao and Zhang, Danyang and Chen, Jixuan and Li, Xiaochuan and others},
  journal = {arXiv preprint arXiv:2404.07972},
  year    = {2024}
}

@article{grpo,
  title = {DeepSeekMath: Pushing the Limits of Mathematical Reasoning in Open Language Models},
  author = {Shao, Zhihong and Wang, Peiyi and Zhu, Qihao and Xu, Runxin and Song, Junxiao and Bi, Xiao and Zhang, Haowei and Zhang, Mingchuan and Li, Y. K. and Wu, Y. and Guo, Daya},
  journal = {arXiv preprint arXiv:2402.03300},
  year = {2024}
}

@inproceedings{lora,
  title = {Lo{RA}: Low-Rank Adaptation of Large Language Models},
  author = {Hu, Edward J. and Shen, Yelong and Wallis, Phillip and Allen-Zhu, Zeyuan and Li, Yuanzhi and Wang, Shean and Chen, Lu and Chen, Weizhu},
  booktitle = {International Conference on Learning Representations},
  year = {2022}
}

@article{taubench,
  title = {$\tau$-bench: A Benchmark for Tool-Agent-User Interaction in Real-World Domains},
  author = {Yao, Shunyu and Shinn, Noah and Razavi, Pedram and Narasimhan, Karthik},
  journal = {arXiv preprint arXiv:2406.12045},
  year = {2024}
}

@inproceedings{her,
  title = {Hindsight Experience Replay},
  author = {Andrychowicz, Marcin and Wolski, Filip and Ray, Alex and Schneider, Jonas and Fong, Rachel and Welinder, Peter and McGrew, Bob and Tobin, Josh and Abbeel, Pieter and Zaremba, Wojciech},
  booktitle = {Advances in Neural Information Processing Systems},
  year = {2017}
}

@article{vpr,
  title = {Verifiable Process Rewards for Agentic Reasoning},
  author = {Yuan, Huining and Xu, Zelai and Wang, Huaijie and Yi, Xiangmin and others},
  journal = {arXiv preprint arXiv:2605.10325},
  year = {2026}
}

@inproceedings{ng1999,
  title     = {Policy Invariance Under Reward Transformations: Theory and Application to Reward Shaping},
  author    = {Ng, Andrew Y. and Harada, Daishi and Russell, Stuart},
  booktitle = {International Conference on Machine Learning (ICML)},
  year      = {1999}
}

@inproceedings{rudder,
  title     = {{RUDDER}: Return Decomposition for Delayed Rewards},
  author    = {Arjona-Medina, Jose A. and Gillhofer, Michael and Widrich, Michael and Unterthiner, Thomas and Brandstetter, Johannes and Hochreiter, Sepp},
  booktitle = {Advances in Neural Information Processing Systems (NeurIPS)},
  year      = {2019}
}

@inproceedings{hca,
  title     = {Hindsight Credit Assignment},
  author    = {Harutyunyan, Anna and Dabney, Will and Mesnard, Thomas and Azar, Mohammad Gheshlaghi and Piot, Bilal and Heess, Nicolas and van Hasselt, Hado and Wayne, Greg and Singh, Satinder and Precup, Doina and Munos, R{\'e}mi},
  booktitle = {Advances in Neural Information Processing Systems (NeurIPS)},
  year      = {2019}
}

@article{ge2026coverage,
  title   = {Coverage, Not Credit: Failure-Credit Routing of Zeroth-Order Perturbation Budgets Does Not Improve On-Pool Sample Efficiency for {LLM} Agents},
  author  = {Ge, Yuxu},
  journal = {arXiv preprint arXiv:2608.28011},
  year    = {2026}
}

\end{document}